\documentclass[letterpaper, 10 pt, conference]{ieeeconf}  %

\IEEEoverridecommandlockouts                              %

\usepackage{cite}
\usepackage{amsmath,amssymb,amsfonts}
\usepackage{algorithmic}
\usepackage{longtable}
\usepackage{float}
\usepackage{graphicx}
\usepackage{textcomp}
\usepackage{tabularx}
\usepackage{adjustbox}
\usepackage{subcaption}
\usepackage{bm}
\usepackage{hhline}
\usepackage{threeparttable}
\usepackage{xcolor}
\usepackage{tablefootnote}
\usepackage{booktabs}
\usepackage{hyperref}

\title{\LARGE \bf
Large Language Models as Zero-Shot Human Models for Human-Robot Interaction
}

\author{Bowen Zhang$^{1}$ and Harold Soh$^{1,2}$%
\thanks{This research is supported by the National Research Foundation Singapore and DSO National Laboratories under the AI Singapore Programme (AISG Award No: AISG2-RP-2020-017).}
\thanks{$^{1}$Dept. of Computer Science, National University of Singapore \texttt{\{bowenzhang, harold\}@comp.nus.edu.sg}. $^{2}$Smart Systems Institute (SSI), NUS.}
}

\newcommand{\prompt}[1]{\textsf{\footnotesize{#1}}}

\begin{document}

\newcommand{\davinci}{\textsc{{DaVinci}}}
\newcommand{\tfive}{\textsc{{T5}}}
\newcommand{\baseline}{\textsc{{baseline}}}
\newcommand{\bfit}[1]{\textit{\textbf{#1}}}

\newcommand{\para}[1]{\vspace{0.5em}\noindent\textbf{#1}}

\maketitle
\thispagestyle{empty}
\pagestyle{empty}
\begin{abstract}
Human models play a crucial role in human-robot interaction (HRI), enabling robots to consider the impact of their actions on people and plan their behavior accordingly. However, crafting good human models is challenging; capturing context-dependent human behavior requires significant prior knowledge and/or large amounts of interaction data, both of which are difficult to obtain. In this work, we explore the potential of large language models (LLMs) --- which have consumed vast amounts of human-generated text data --- to act as zero-shot human models for HRI. Our experiments on three social datasets yield promising results; the LLMs are able to achieve performance comparable to purpose-built models. That said, we also discuss current limitations, such as sensitivity to prompts and spatial/numerical reasoning mishaps.
Based on our findings, we demonstrate how LLM-based human models can be integrated into a social robot's planning process and applied in HRI scenarios focused on the important element of trust. Specifically, we present one case study on a simulated trust-based table-clearing task and replicate past results that relied on custom models. Next, we conduct a new robot utensil-passing experiment ($n = 65$) where preliminary results show that planning with an LLM-based human model can achieve gains over a basic myopic plan. In summary, our results show that LLMs offer a promising (but incomplete) approach to human modeling for HRI. %
\end{abstract}

\section{Introduction}

Human models endow robots with the ability to reason about humans, enabling human-centered and personalized robot behavior~\cite{choudhury2019utility}. Despite the advantages, accurate human modeling remains a significant challenge. Handcrafted human models typically encode strong assumptions (e.g.~\cite{ziebart2009planning, sadigh2016planning, baker2017rational, chen2018planning}), which can limit flexibility and are challenging to scale up to real-world settings. An alternative approach is to use non-parametric data-driven models(e.g.~\cite{chen2022mirror, schmerling2018multimodal}). These models have become popular of late due to the success of deep learning in a variety of fields, from computer vision to game-playing. However, these methods typically require large amounts of human interaction data that is difficult and costly to collect.  

In this work, we attempt to circumvent these issues by using large pre-trained or ``foundation'' models. Specifically, we examine large language models (LLMs) that have caused excitement due to their impressive abilities to generate plausible human-sounding text and their performance on downstream tasks that they were not explicitly trained for. Here, we ask: \textbf{can LLMs function effectively as human models for HRI?}

\begin{figure}
    \centering
    \includegraphics[width=\columnwidth]{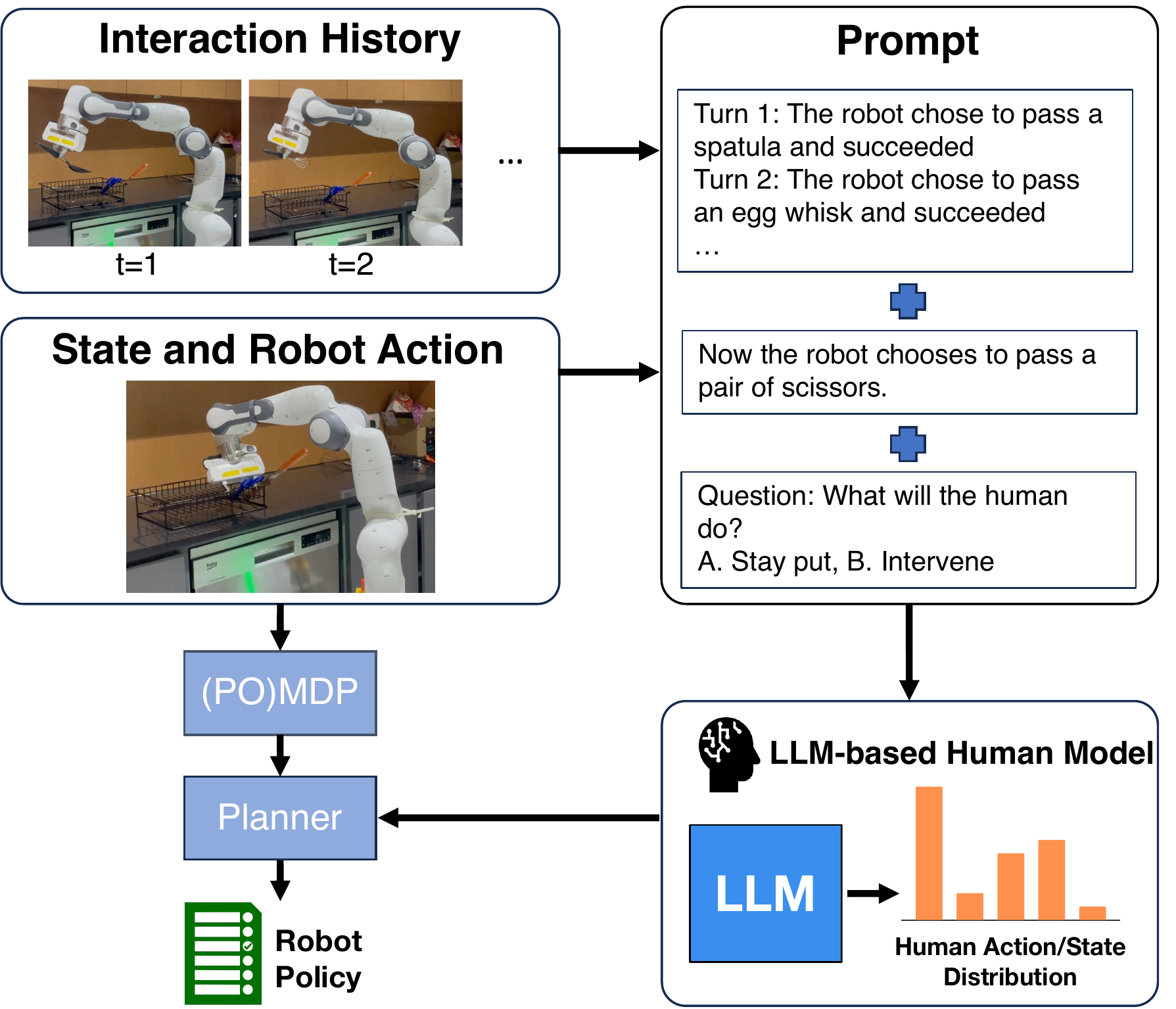}
    \caption{In this work, we explore how large language models (LLMs) can be used as zero-shot human models in HRI. We first evaluate the effectiveness of such models using benchmark datasets. Then, we demonstrate how LLM-based models can be used in planning for two trust-related HRI scenarios.}
    \label{fig:mainfig}
    \vspace{-10pt}
\end{figure}

Since LLMs have been trained on vast amounts of human-generated text, it is conceivable that they have captured elements of human behavior. 
If the answer to our question turns out in the affirmative, LLMs would be a boon for the HRI community, particularly those of us developing/using computational models for human-robot collaboration and interaction~\cite{thomaz2016computational,nocentini2019survey}. 
We are especially interested in the LLMs' \emph{zero-shot} capabilities since no further human data is required. 
On the other hand, there are good reasons to expect LLMs to be poor substitutes for existing human models --- HRI tasks involve both social and physical aspects, and recent evidence indicates that LLMs can fail spectacularly on tasks that require skills such as physical reasoning that go beyond linguistic competence (e.g.~\cite{kojima2022large, valmeekam2022large, huang2022language, yaqi23translating}).  

We first present an empirical study that shows that LLMs indeed well-capture human latent states and behavior. Using three different datasets, we show that two state-of-the-art LLMs (FLAN-T5 and a variant of GPT-3.5) can achieve predictive performance comparable to specialized machine learning models. This was achieved using a zero-shot approach, \emph{without the use of any additional training data}. While it is tempting to conclude that LLMs are effective human models, the situation is more complex. A deeper analysis of our results shows that the LLMs underperformed on tasks that require spatial and numerical reasoning and are sensitive to prompt syntax. This can make application in real-world HRI scenarios challenging and suggest that LLMs may be best used as ``task-level'' human models.

To take a step forward, we explore how LLM-based human models can be used in a planning setup for HRI (Fig.~\ref{fig:mainfig}). We first experiment with a simulated table-clearing setup used in prior work on human-robot trust~\cite{chen2018planning} and achieve comparable results. We then discuss preliminary results from a new utensil-passing experiment with $n=65$ participants. In this more complex setup, a planner equipped with an LLM-based human model generated a reasonable plan that mitigated over-trust. Our experiment also revealed that some participants based their judgments on the physical characteristics of objects, which our model did not account for. This suggests the complexity of HRI tasks may require human models that combine LLMs with ``lower-level'' models that consider geometry, motion, and other physical characteristics, which would make interesting future work.

\section{Preliminaries And Related Works}

\para{Large Language Models.}
A Language Model (LM) is a distribution over sequences of word tokens. Autoregressive LMs model the conditional distribution $p(w_k|w_{<k})$ over $w_k$, the next token at the $k^{th}$ position, given a sequence of previous input tokens $w_{<k}$. 
State-of-the-art LLMs have tens or hundreds of billions of parameters, e.g. OpenAI's GPT-3~\cite{brown2020language} and Google's PaLM~\cite{chowdhery2022palm}). When trained with massive amounts of text data in a self-supervised manner, these LLMs are not only capable of generating human-like text but also demonstrate impressive performance on downstream tasks. 

In this work, we focus on the ability of LLMs to generate plausible human text that is consistent with prior information fed in as a sequence of word tokens called a \emph{prompt}. Our work is related to very recent research on replicating human behavior on classic psychological/economic experiments with LLMs~\cite{aher2022using} and the possible emergence of theory-of-mind in LLMs~\cite{kosinski2023theory}, which remains highly debatable~\cite{ullman2023large}. The key difference is that we focus on the applicability of LLMs to HRI, which involves scenarios that include a robot and  physical-social interaction. 
Another related thread of work is in planning with language models. Prior work has used LLMs to predict object affordance~\cite{ahn2022can} or to directly generate action plans~\cite{huang2022language}. However, recent work has also shown that LLMs can be poor planners~\cite{valmeekam2022large}. Our work differs from the above as we propose to use LLMs to approximate human behavior and use the LLM-based human models for planning to obtain a suitable robot policy.

\para{Human Models for HRI.}
In general, modeling humans is a large interdisciplinary endeavor and here, we focus on computational models for HRI. We briefly review recent work and refer readers wanting more comprehensive coverage to excellent survey articles~\cite{thomaz2016computational,choudhury2019utility}. 

Existing approaches to modeling human behavior in HRI can be broadly categorized into two paradigms: Theory-of-Mind (ToM) models and `black-box' data-driven models. ToM models broadly refer to methods that incorporate a set of assumptions about human mental processing and behavior. For example, Bayesian Theory of Mind (BToM) assumes humans behave rationally and update their beliefs in a Bayesian manner~\cite{baker2017rational,lee2019bayesian, baker2014modeling}. %
In contrast, `black-box' models make few assumptions and model human behavior in a data-driven manner (e.g.~\cite{schmerling2018multimodal}). Such methods typically require large amounts of real human data, which can be difficult to obtain. Recent approaches have proposed training agents in a multi-agent manner, e.g., Fictitious Co-Play~\cite{strouse2021collaborating}, which obviates the need for human data. However, these RL-based methods are computationally expensive to train and require significant amounts of interaction data that may be  costly or impossible to acquire in the real world. A recent hybrid method, MIRROR~\cite{chen2022mirror}, models a human based on a robot's internal self-model (trained with RL) and then uses a small amount of human data to adapt the model to a particular individual. However, it is unclear if MIRROR or any of the above methods can well capture human mental states, e.g., emotions, which are crucial in many real-world settings. Here, we explore whether LLM-based human models can overcome the above shortcomings.

\section{LLMs as Zero-Shot Human Models}

\subsection{Problem Statement}

Consider an environment that is populated by $N$ agents $a^{(i)} \in \mathcal{A}$ for $i=1, \dots, N$. An agent may be a human or a robot. Given an environment state $s_t \in \mathcal{S}$ at time $t$ and a history of interactions $h_t = \{ (s_k, \{u^{(i)}_{k}\}_i) \}_{k=1}^{t-1} \in \mathcal{H}$ 
where $s_k$ represents the environment state at time $k$ and $u^{(i)}_k$ represents the action of agent $i$ at time $k$, we aim to model a specific property $z_t^H \in \mathcal{Z}^H$ of one of the human agents $a^H \in \mathcal{A}$. More precisely, $z_t^H$ is an abstract scenario-specific random variable (e.g., a human action or mental state) and we seek to model the distribution $p(z^H_t|s_t, h_{t})$. For example, $z^H_t$ may be the human's trust in the robot to perform a specific task.

We assume that the environment state, agents, and history can be textualized via a function $f$ to a \emph{prompt}, $x_{t} = f(s_t, h_{t}, \mathcal{A})$. We denote the LLM-based human model as $p_{l}(z_t^H|x_t)$ where for simplicity, we assume that the LLM can output a distribution over $z_t^H$ or that a suitable mapping can be defined. Our study aims to evaluate if $p_{l}(z_t^H|x_t)$ well approximates $p(z^H_t|s_t, h_{t})$.

\begin{figure*}
    \begin{center}
    \includegraphics[width=1.0\textwidth]{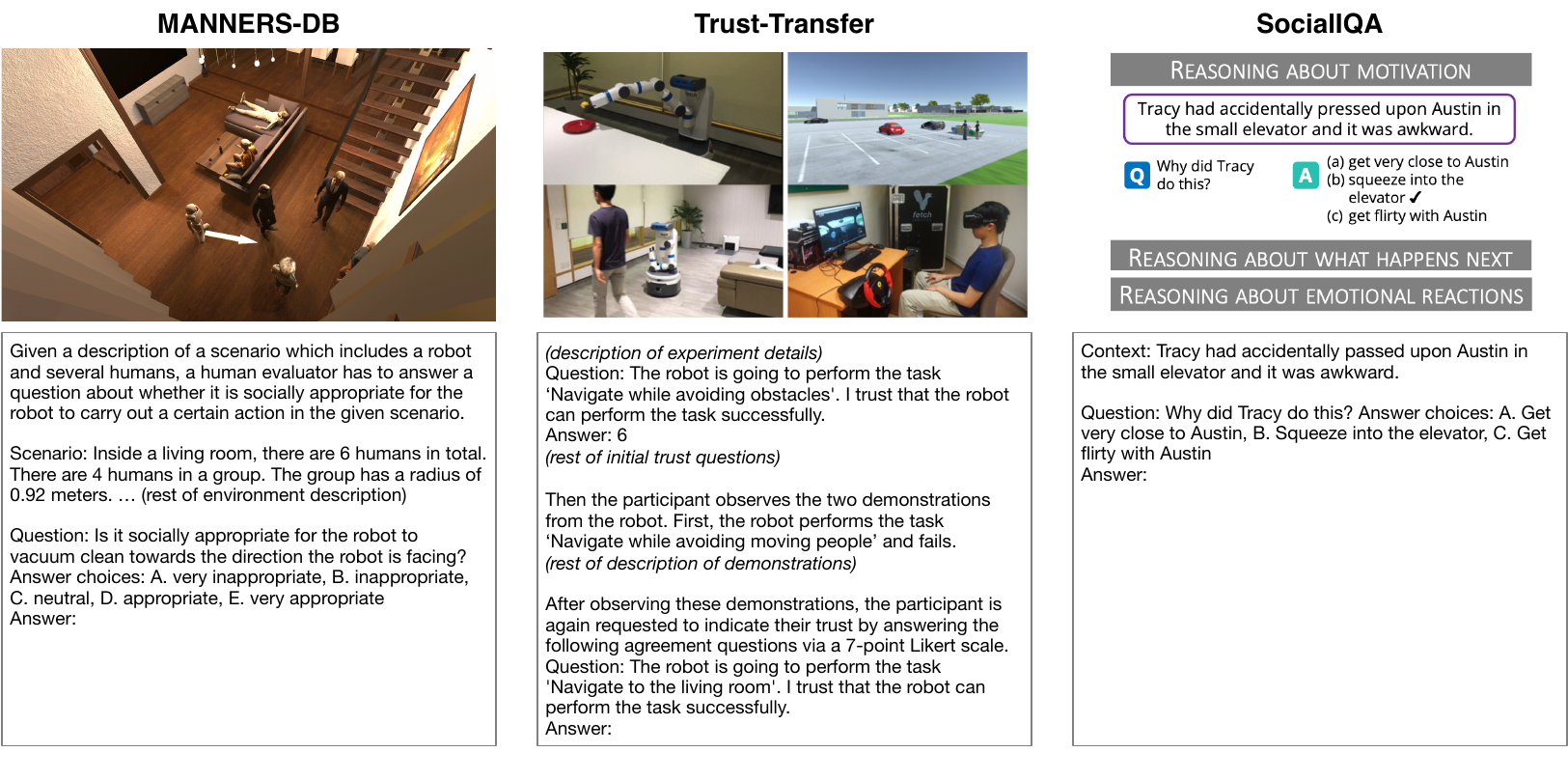}
    \end{center}
    \caption{Datasets and Example Prompts in Prediction Experiments. We use two HRI datasets: MANNERS-DB~\cite{tjomsland2022mind} and Trust-Transfer~\cite{soh2018transfer,soh2020multi}, and included SocialIQA~\cite{sap2019socialiqa}, a general social reasoning benchmark for human interactions. For each dataset, we show an illustrative image (reproduced from ~\cite{tjomsland2022mind,soh2020multi,sap2019socialiqa}, respectively) and an example prompt.}
    \label{fig:datasets_prompts}
\end{figure*}

\subsection{Models, Datasets, and Prompts}

\para{Models.} We work with two different LLMs: (i) text-davinci-003 (175B)~\cite{ouyang2022training} and (ii) FLAN-T5-XXL (11B)~\cite{chung2022scaling}, which we will refer to as \davinci{} and \tfive{}, respectively. \davinci{} a large model that is finetuned with human feedback and achieves state-of-the-art performance. \tfive{} is a smaller open-source LLM optimized with instruction fine-tuning. For our experiments, we ran \tfive{} on a local workstation and accessed \davinci{} using the OpenAI API. Both LLMs can either provide samples or a distribution over next tokens.

\para{Datasets.} We use two HRI-related datasets:
\begin{itemize}
    \item \textbf{MANNERS-DB}~\cite{tjomsland2022mind}, a recent dataset for assessing the appropriateness of robot actions in specific social contexts. The test set comprises 800 instances, where each instance is a simulated household scenario and a description of a robot action. The scenarios are divided into two sets based on whether the robot actions are executed within a circle surrounding itself or in a direction indicated by an arrow. Each instance is labeled by 15 human annotators, who are presented with an image of a simulated scene and asked to rate the degree of social appropriateness of the robot action from 1 (very inappropriate) to 5 (very appropriate). %
    This dataset enables us to evaluate whether LLM-based human models can capture the effects of physical and spatial features in HRI. 
    \item \textbf{Trust-Transfer}~\cite{soh2018transfer} is a dataset on how human trust in a robot's capability changes after observing it perform tasks. The dataset encompasses two different domains (household and driving) and contains 189 instances. Each instance includes a participant's initial level of trust on three \emph{test tasks}, two observed demonstrations, and the participant's final level of trust on the same three test tasks \emph{after} observing the demonstrations. Participants were asked to indicate their level of trust in the robot using a 7-point Likert scale. We included the Trust-Transfer dataset (i) to evaluate whether LLMs can capture the effect of interaction history on a human's mental state, and (ii) because the dataset is closely related to our trust-based case studies in Sec.~\ref{sec:planning}.
\end{itemize}
and a general human-human social interaction dataset:
\begin{itemize}
    \item \textbf{SocialIQA}~\cite{sap2019socialiqa}, a benchmark designed to measure the ability to perform social commonsense reasoning. The test split contains 1954 instances, each comprising a context, a question, and three answer choices with one ground truth label. Performing well on SocialIQA would demonstrate LLM's ability to reason about human social interactions, including intent. For example, a robot could be observing an interaction between two or more humans and has to infer their mental states.
\end{itemize}

\para{Prompt Design.} The precise prompts used varied between datasets (see Fig.~\ref{fig:datasets_prompts} for examples) but in general, we follow the practice suggested by~\cite{aher2022using} to craft clear templated prompts that minimize the invalid completions, e.g. the probability of generating anything other than the listed choices when given a multiple-choice question.
The prompts were designed to include:
\begin{itemize}
    \item Information about the experiment that was provided to human participants in the original HRI/social experiment, e.g., rules of the task and features of the robot;
    \item The interaction history, current environment, and a description of the robot $a^R$'s action (in the HRI datasets);
    \item A query to probe LLM's response. To extract a probability distribution over $z^H_t$, we frame our prompts such that the number of valid completions is limited to elements in $Z^H$; we assign a single-token label to each $z^H$ via a multiple choice question or a Likert scale question\footnote{The LLMs occasionally generated invalid completions ($<3\%$ of the instances in our experiments). These are placed in a single ``catch-all'' class and marked as errors.}. As such, the length of each valid completion is a single token and we avoid favoring short answers when comparing token probabilities. 
\end{itemize}
Due to space restrictions, we provide specific details on the prompts for each dataset in an online supplementary material~\cite{suppmat}.

\subsection{Baseline Models, Evaluation Criteria, and Limitations}

\para{Baseline Models.} We compared the LLMs to specialized published models created for each of the datasets:
\begin{itemize}
    \item \textbf{BNN-16CL}\footnote{We ran BNN-16CL using code provided by the authors at  \url{https://github.com/jonastjoms/MANNERS-DB}. Our reported numbers are different from~\cite{tjomsland2022mind} possibly due to a difference in the test dataset split.}~\cite{tjomsland2022mind}, a Bayesian continual learning model trained on the MANNERS-DB dataset to  predict social appropriateness of robot actions.
    \item \textbf{GPNN}, a hybrid Gaussian Process and Neural Network model developed for predicting Trust-Transfer and applied to the Trust-Transfer dataset.
    \item \textbf{BERT-large}~\cite{devlin2018bert}, a 340M-parameter language model fine-tuned as a classifier for the SocialIQA dataset.
\end{itemize}
When reporting results, we use the umbrella term \baseline{} model to refer to the {respective} model for each dataset.

\para{Evaluation Measures.} We primarily evaluated the above models using an error score and a Consistency with Mode (CwM) score. The error score captures how well a given model's output coincides with the dataset targets on average. This score is \emph{dataset-dependent} and refers to the metric used by the authors for each dataset; this allowed us to directly compare against previously reported numbers (when available). The error scores are RMSE (root mean square error), MAE (mean absolute error), and error rate for {MANNERS-DB}, {Trust-Transfer} and {SocialIQA}, respectively. For simplicity, we will refer to these different measures as simply the ``error score''. %

The CwM score measures the consistency between the LLMs and the majority human label. To produce a consistent score across the different datasets, we first binarize the targets; for MANNERS-DB, if a majority of annotators rated the appropriateness level of a robot action as $\geq$ 3 (neutral), we labeled it as socially acceptable (otherwise, it was labeled as not socially acceptable). An analogous operation was performed for the Trust-Transfer dataset for scores $\geq$ 4. The SocialIQA dataset only provides a single label per instance and we took this as the mode. 

To give us a sense how well the distribution $p_l(z_t^H|x_t)$ matches $p(z_t^H|s_t,h_{t-1})$, we include two additional measures: (i) an entropy relative similarity score $ S(q, p) = \frac{\sum_{z^H}q(z^H)\log q(z^H)}{\sum_{z^H}p(z^H)\log p(z^H)} $, and (ii) the 2-Wasserstein metric $W_2(q, p)$. We set $q$ to be the model output and $p$ to be the target empirical distribution.
Intuitively, the entropy of a distribution captures how ``broad'' the distribution is or its inherent uncertainty. The $S(q, p)$ score reveals whether a model output distribution $q$ captures the spread of the target $p$ or if it is under/over-confident --- a value of 1 indicates the same entropy, while $S(q,p) < 1$ indicates a narrower spread (and vice-versa). The 2-Wasserstein metric captures how similar two distributions are; it measures how much probability mass has to be moved to transform $q$ to $p$. 
Note that these two distribution-related measures could only be computed for MANNERS-DB (since Trust-Transfer and SocialIQA only contain one label per instance).

\para{Limitations.} For both MANNERS-DB and SocialIQA, the labels are provided by third-party humans who are not actually part of the environment. Based on prior work~\cite{heyes2014cultural, zhang2012perspective}, we assume that human annotators are able to provide sound assessments of another person's mental state or behavior. 
Next, reliance on handcrafted prompts is an important limitation, better performance may be possible with alternative prompt structures. Finally, an issue when evaluating LLMs is the risk of ``data leakage'', i.e., a given dataset has been seen by the LLM during training. Since training datasets for both \davinci{} and \tfive{} are not readily available, we are unable to confirm whether this was the case. That said, data leakage for SocialIQA is unlikely~\cite{srivastava2022beyond} and the HRI datasets (prompts and answers) were created for this study and hence, have not been \emph{directly} seen by the LLMs. However, the raw datasets on which our prompts are based are publicly available and LLMs may have either seen the raw data or relevant papers. We attempt to mitigate this by extending the trust task with items that were not in the original dataset~\cite{soh2020multi} in Sec.~\ref{sec:planningexp2}.

\begingroup

\begin{table*}\centering
    \caption{Error and CwM Scores and Distribution Measures. $S(q, p)$ closer to 1 indicates better capture of the spread of human distribution.}\label{tab:summary_error_metrics_distribution_measures}
\begin{threeparttable}
\begin{adjustbox}{width=\textwidth}
\begin{tabular}{ll|ccc|ccc|ccc|ccc}
\hline\hline
& & \multicolumn{3}{c|}{\textbf{Error Score} ($\downarrow$)} & \multicolumn{3}{c|}{\textbf{CwM} ($\uparrow$)} & \multicolumn{3}{c|}{\bm{$S(q, p)$}} & \multicolumn{3}{c}{\bm{$W_2(q, p)$} ($\downarrow$)}\\
\textbf{Dataset} & \textbf{Context} & \davinci{} &\tfive{}& \baseline & \davinci{} & \tfive{} & \textsc{Human} & \davinci{} &\tfive{}& \baseline & \davinci{} &\tfive{}& \baseline \\
\hline
MANNERS &Circle &0.787 &\textbf{0.708} &0.876 &0.730 &0.750 &\textbf{0.760} & 0.467 &1.354 & \textbf{1.127} & 0.168 &\textbf{0.164} & 0.173 \\
&Arrow &\textbf{0.695} &0.762 &0.664 &\textbf{0.734} &0.632 &0.717 & 0.586 &1.226 & \textbf{1.199} & 0.206  &0.141 & \textbf{0.124}\\
&Overall &\textbf{0.737} &0.739 &0.764 &0.733 &0.684 &\textbf{0.736} & 0.645 & 1.281 & \textbf{1.159} & 0.189 &0.151 & \textbf{0.145} \\
\hline
Trust-Transfer &Household &0.162 & 0.211 &\textbf{0.156} &\textbf{0.896} &0.760 &- &- &- &- &- &- &- \\
    &Driving &\textbf{0.154} & 0.211 &0.163 &\textbf{0.806} &0.720 &- &- &- &- &- &- &- \\
&Overall &\textbf{0.158} &0.211 &0.159 &\textbf{0.852} &0.741 &- &- &- &- &- &- &- \\
\hline
SocialIQA & Overall &0.278 &\textbf{0.181} &0.340 &0.722 & 0.819 & 0.684\tnote{1} &- &- &- &- &- &-
\\
 &  & & &  &  & & \textbf{0.869}\tnote{2} &- &- &- &- &- &-
\\
\hline
\hline
\end{tabular}
\end{adjustbox}
\begin{tablenotes}
\item[1] Average human performance, reported in \url{https://github.com/google/BIG-bench}.
\item[2] Best human performance, reported in\cite{sap2019socialiqa}.
\end{tablenotes}
\end{threeparttable}
\vspace{-10pt}
\end{table*}
\endgroup

\subsection{Results}

The error and CwM scores are summarized in Table~\ref{tab:summary_error_metrics_distribution_measures}. Across all three datasets, \davinci{} achieves surprisingly good zero-shot performance, sometimes achieving lower error scores than the specialized models. \tfive{} also achieves good scores on the MANNERS-DB and SocialIQA datasets but performs poorly on the Trust-Transfer datasets; we delve into the reasons for this subpar performance in later analysis. The CwM scores indicate that the LLMs achieve human-level consistency to the majority label on the MANNERS-DB and SocialIQA datasets. We observe disagreement between human annotators in the SocialIQA dataset since the average annotator is only 68.4\% consistent with the majority.

Table~\ref{tab:summary_error_metrics_distribution_measures} also shows the distribution measures on MANNERS-DB. In general, the baseline (a Bayesian model) best captures the target distribution. 
\davinci{} tends to under-capture the spread of the distribution, as indicated by its lower entropy similarity $S(q,p)$ scores. 
In contrast, \tfive{} has scores $>1$, suggesting an overly broad distribution, but one that better matches the empirical human label distribution; the Wasserstein metric is consistently lower for \tfive{} compared to \davinci{} in all contexts. 

Taken together, the above results indicate that the LLMs achieve performance comparable to strong baseline models. We emphasize that this was achieved in a \emph{zero-shot} manner, suggesting that \textbf{LLMs can be effective zero-shot human models for HRI without further training or fine-tuning}.

\subsection{Discussion and Analysis on HRI Datasets} 
\label{subsec:disc_hri}

The results above are promising, but the scores also show that the LLMs are imperfect models, e.g., \tfive{} performs poorly on the Trust-Transfer dataset. Next, we provide further analysis into the errors made by the LLMs to highlight pitfalls and potential improvements.

\begin{table}
\begin{center}
\caption{CwM Scores obtained by \davinci{} for each individual Action Type in MANNERS-DB.}
\label{tab:manners_classification_results_davinci}
\begin{tabular}{l c c c}
\hline
\hline 
\textbf{Robot Action} & \textbf{CwM} ($\uparrow$) & \textbf{Precision} ($\uparrow$) & \textbf{Recall} ($\uparrow$)  \\
\hline
Vacuum cleaning             & 0.560 & 0.506 & 0.933\\
Mopping the floor           & 0.520 & 0.478 & 1.000\\
Carry warm food             & 0.740 & 0.875 & 0.814\\
Carry cold food             & 0.670 & 0.944 & 0.698\\
Carry drinks                & 0.940 & 0.959 & 0.979\\
Carry small objects         & 0.960 & 0.960 & 1.000\\
Carry big objects           & 0.710 & 0.583 & 0.600\\
Cleaning (Picking up items) & 0.886 & 0.886 & 1.000\\
Starting conversation       & 0.661 & 0.755 & 0.841\\
\hline
\hline
\end{tabular}
\vspace{-10pt}
\end{center}
\end{table}

\para{LLMs can perform poorly on HRI tasks that require spatial/physical/numerical reasoning.}
This issue can be seen via a breakdown of the LLM's performance on MANNERS-DB. Table~\ref{tab:manners_classification_results_davinci} shows the CwM scores for \davinci{}, along with precision and recall, on specific actions in the dataset. \tfive{} scores are similar and we relegate the \tfive{} CwM scores (along with a table comparing RMSE values across the models) to the online appendix. Relatively low scores were obtained for `Vacuum cleaning' and `Mopping the floor', which are the two most intrusive actions in the dataset and require an understanding of personal space~\cite{tjomsland2022mind}, which requires physical/spatial and numerical reasoning given the prompt  --- tasks that are known to be difficult for LLMs~\cite{valmeekam2022large,yaqi23translating}. We also observe lower performance on `Carry big objects' and `Starting a conversation', which are also actions where social appropriateness found to be highly correlated with inter-agent distances~\cite{tjomsland2022mind}. %

To further investigate this issue, we examined the 89 failure cases of `Vacuum cleaning' and `Mopping the floor'. Through manual inspection of the scene images and annotators' comments, we identified 56 failure scenarios where the robot's action clearly invades a human's personal space, e.g., a human is standing in the robot's intended working area, but the LLM predicts that the action is socially appropriate. For these instances, we directly queried the LLM with a Yes-No question:  ``\textsf{\small Does the robot's designated working area intrude on anybody's personal space?}". Both \davinci{} and \tfive{} gave wrong answers to all 56 questions. We also attempted Zero-Shot Chain-of-Thought prompting on \davinci{}, which has been shown to elicit more reasonable responses~\cite{kojima2022large}. Performance improved but remained unsatisfactory (15/56 correct, $26.7\%$). An instructive failure response is as follows: 
\begin{quote}
    ``\textsf{\small First, the closest human is 0.53 meters away from the robot. This means that the robot's designated working area of 2.00 meters will not intrude the closest human's personal space.}''
\end{quote}
We observed that the response is linguistically correct and leverages information in the prompt, but the drawn conclusion is incorrect since the human is present in the cleaning area.  %

\begin{figure}
\centering
\fbox{
\parbox{0.95\columnwidth}{%
\prompt{
... \textit{(description of experiment details)}\newline
The participant rates their trust on the task 'Pick and place a plastic can' as 5 out of 7. \newline
... \textit{(rest of initial trust descriptions and description of demonstrations)}\newline
Given these demonstrations and the initial trust, now the participant will rate their trust on the task `Pick and place a plastic can' as 
}
}
}
\caption{Example altered prompt used in trust-transfer experiment in the household domain.}
\label{fig:trust_transfer_prompts_alter}
\vspace{-10pt}
\end{figure}
\begin{table}
\caption{Results on Trust-Transfer using Altered Prompts.}
\begin{center}
\begin{tabular}{l c c c c c c c c}
\hline
\hline
\textbf{}&\multicolumn{3}{c}{\textbf{Error Score($\downarrow$)}} & \multicolumn{2}{c}{\textbf{CwM($\uparrow$)}}\\

\textbf{Domain} & \davinci{} & \tfive{} & \baseline{} & 
\davinci{} & \tfive{}\\
\hline
Household & 0.163 & 0.170 & 0.156 & 0.875 & 0.796 \\

Driving & 0.155 & 0.160 & 0.163 & 0.806 & 0.802 \\

Overall & 0.159 & 0.165 & 0.159 & 0.841 & 0.799 \\
\hline
\hline
\end{tabular}
\end{center}
\vspace{-10pt}
\label{tab:trust_transfer_result_alter}
\end{table}

\para{LLMs are sensitive to the prompt structure.} 
Recall that \tfive{} displayed subpar performance on Trust-Transfer. An examination of \tfive{}'s outputs revealed that it had a strong tendency to predict participants' post-observation trust to be the same as their initial trust. We hypothesized that this was caused by our prompt resembling the structure of few-shot prompts, which biased the model into copying the initial answers since the questions were identical (see Fig.~\ref{fig:datasets_prompts}). 
To test our conjecture, we designed an alternative prompt that altered how the initial trust and the query are presented (Fig~\ref{fig:trust_transfer_prompts_alter}). Table~\ref{tab:trust_transfer_result_alter} shows that the new prompt style significantly improves \tfive{}'s performance. 

\para{Summary.} Our study shows that LLMs are surprisingly adept at functioning as zero-shot human models. That said, our analysis also reveals that (i) LLMs can perform poorly on HRI tasks that require spatial and numerical understanding and (ii) LLMs are sensitive to prompt design. 
These findings add to the growing literature on the strengths and limitations of LLMs~\cite{valmeekam2022large,yaqi23translating,ullman2023large,kosinski2023theory,aher2022using}. Within the context of HRI, care should be taken in the application of LLMs; HRI is typically grounded in the real world and thus, involves spatial and physical reasoning. Overall, our results suggest that \textbf{LLMs are better-suited as task-level (symbolic) human models, and alternative ``low-level'' models may be needed to account for geometry and motions in a continuous space}.

\section{Case-Studies: Planning with LLM-based human models}
\label{sec:planning}

In this section, we move from prediction to planning for human-robot interaction. Leveraging insights gained from our study, we demonstrate how an LLM-based human model can be used in a planner for HRI. We begin by replicating a prior study on trust in a table-clearing task and then move on to a new utensil-passing experiment.

\subsection{Problem Statement}
Given our previous findings, our HRI scenarios are framed at the task level and formalized as MDPs. At each time-step $t$, a world state $s_t$ is fully observed by both the human $H$ and the robot $R$. The robot takes an action $u^R_t$ and in response, the human takes an action $u^H_t$. The world state then changes according to a transition function $p(s_{t+1}|s_t,u^H_t,u^R_t)$ and a reward $r(s_{t},u_t^H,u_t^R,s_{t+1})$ is received. The robot's optimal policy $\pi^R_*$ is therefore,
\begin{align*}
\pi^R_* = \underset{\pi_R}{\arg\max} \mathop{\mathbb{E}}_{u^H_t \sim \pi^H, u^R_t \sim \pi^R} \sum_{t=0}^{\infty} \gamma^t r(s_t, u^H_t, u^R_t, s_{t+1})
\end{align*}
where $\gamma$ is the discount rate. We approximate the unknown human policy $\pi^H$ with $p^{l}(z^H_t|x_t)$ where $z^H_t = u^H_t$ and $x_t = f(s_t, h_t, u^R_t, \{H,R \})$ and assume known transition dynamics. As such, the robot decision process becomes a MDP and off-the-shelf planners can be used. In the following experiments, we use simple offline value iteration since the MDPs are relatively small with short planning horizons. Future work can investigate more complex scenarios using more sophisticated planners and LLM-based models.

\subsection{Table-clearing experiment}

\para{Experiment Setup.}
We first test our LLM-based planner on a simulated table-clearing experiment~\cite{chen2018planning}, where a human and a robot collaborate to clear objects off a table. The objects include three water bottles, one fish can, and one wine glass. At each time step, the robot chooses one of the objects to remove. The human then chooses whether to intervene and pick up the object, or stay put and let the robot remove the object. If the human stays put and the robot succeeds, they will get a reward based on the object: 1 for plastic bottle, 2 for fish can, and 3 for wine glass. However, if they stay put and the robot fails, they will receive a penalty: no penalty for plastic bottle, 4 for fish can and 9 for wine glass. If they choose to intervene, they will receive no reward or penalty. It is assumed that the robot will never fail but this information is not revealed to the human participant.

\para{Prompt Design.} As previously discussed in Sec. \ref{subsec:disc_hri}, prompt structure is important for eliciting correct responses from LLMs. Here, we consider a standard prompt with variations (example in  Fig.~\ref{fig:table_clearing_prompt}): 
\begin{itemize}
    \item \textsc{TC}: We explicitly include the Trust Change in each turn (the most likely post-observation trust change predicted by the LLM model) using a multiple-choice question with options \{increased, decreased, unchanged\} 
    \item \textsc{YN}: Instead of asking which action the human will take when the robot chooses to remove an object, the prompt asks a Yes-No question about whether the human will trust the robot to do so. We assume a deterministic relationship between trust and human action, i.e., the human will intervene if they do not trust the robot to perform the task and stay put otherwise.
\end{itemize}

\para{Results.}
We compare against the \textsc{trust-POMDP} planner~\cite{chen2018planning} using the authors' simulation code. The results of $10^4$ simulation runs are shown in Table~\ref{tab:table_clearing_result}. All \davinci{}-based planners are consistently better than their \tfive{} counterparts and achieve comparable performance with the \textsc{trust-POMDP} planner. Using both \textsc{TC} and \textsc{YN} prompting improved the performance for \tfive{} and \davinci{}-based planners.

\begin{figure}
\centering

\fbox{
\parbox{0.95\columnwidth}{%
\prompt{
\newline
... \textit{(description of experiment setup and rules)} \newline
Turn 1: Robot choice: plastic bottle; Human choice: stay put; Outcome: the robot successfully removes the plastic bottle. \textit{(Include ``The human's trust in the robot increased." in the case of \textsc{TC})} \newline
... \textit{(rest of interaction history)}\newline
\newline
Question: Now the robot chooses to remove the wine glass, what will the human do? Answer choices: A. intervene, B. stay put.
\newline
\textit{OR in the case of \textsc{YN}:} \newline
Question: Will the human trust the robot to remove the wine glass now? Answer choices: A. Yes, B. No.
}
}
}
\caption{Example prompt used in table-clearing experiment.}
\label{fig:table_clearing_prompt}
\vspace{-10pt}
\end{figure}

\begin{table}
\begin{center}
\caption{Simulated Table-clearing Experiment Results.}
\label{tab:table_clearing_result}
\begin{tabular}{l c c c}
\hline
\hline
\textbf{} & \textbf{Mean Return ($\uparrow$)} & \textbf{Interv. prob. on Glass ($\downarrow$)} \\
\cline{1-3} 
\textsc{davinci-TC-YN} & \textbf{6.17 (0.034)} & \textbf{0.352} \\

\textsc{davinci-YN} & 6.13 (0.034) & 0.366 \\

\textsc{davinci-TC} & 6.15 (0.034) & 0.357 \\

\textsc{davinci} & 6.14 (0.034) & 0.360 \\
\hline 
\textsc{T5-TC-YN} & 6.10 (0.034) & 0.368 \\

\textsc{T5-YN} & 6.01 (0.035) & 0.398 \\

\textsc{T5-TC} & 5.94 (0.034) & 0.395 \\

\textsc{T5} & 5.95 (0.035) & 0.405 \\
\hline
\textsc{trust-POMDP} & \textbf{6.17 (0.034)} & \textbf{0.352} \\
\hline
\hline
\end{tabular}
\vspace{-10pt}
\end{center}
\end{table}

\subsection{Utensil-passing Experiment}
\label{sec:planningexp2}

\begin{figure}
\centering
\begin{subfigure}{0.45\columnwidth}
     \centering
     \includegraphics[width=\textwidth]{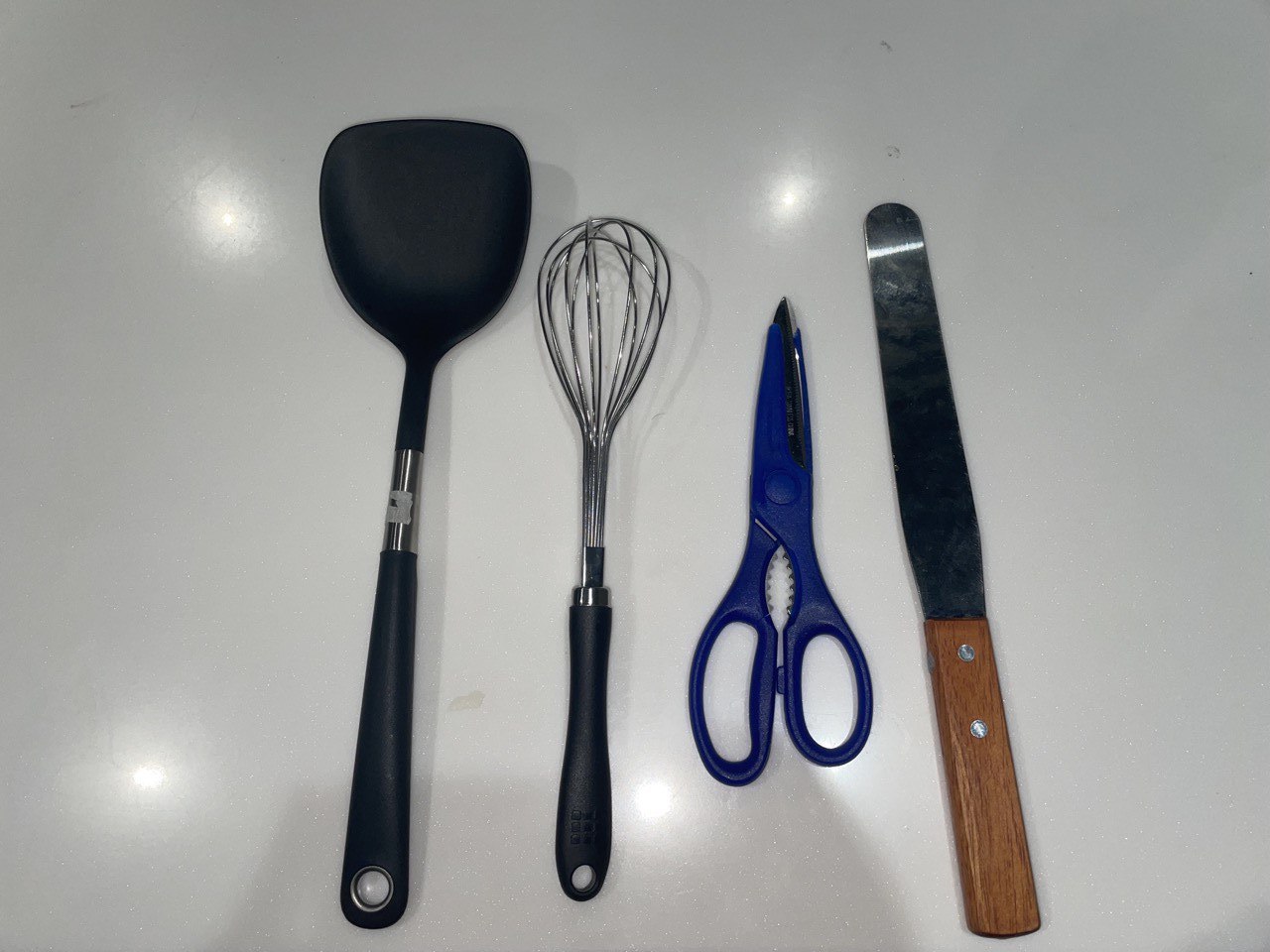}
 \end{subfigure}
 \begin{subfigure}{0.45\columnwidth}
     \centering
     \includegraphics[width=\textwidth]{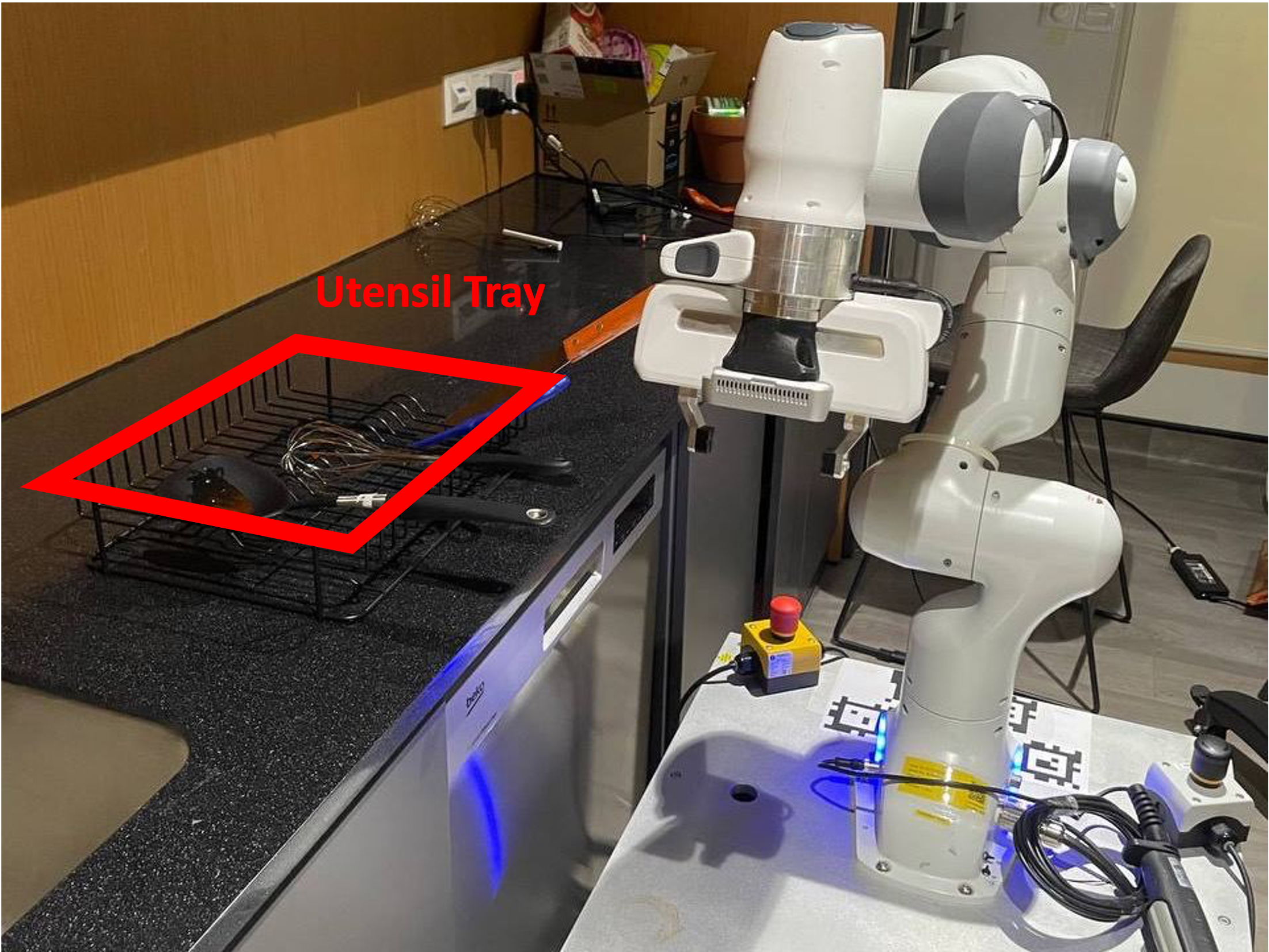}
 \end{subfigure}
 \caption{(Left) Utensils used for the experiment: spatula, egg whisk, scissors and knife. (Right) The experiment environment emulates a kitchen. Utensil tray is highlighted for better visibility.}
\label{fig:utensil_passing}
\vspace{-10pt}
\end{figure}

\para{Experiment Setup.} In this scenario, a human is washing utensils in a kitchen and a robot (a Franka Emika Panda robot arm) is helping to pass dirty utensils to them. The objects include a spatula, an egg whisk, a pair of scissors, and a knife (Fig.~\ref{fig:utensil_passing}). At each time step, the robot chooses one of the objects to pass. The human then chooses between two actions: (A) intervene and retrieve the object by themselves or (B) stay put and let the robot pass it. 

\begin{figure}
    \centering
\includegraphics[width=0.75\columnwidth]{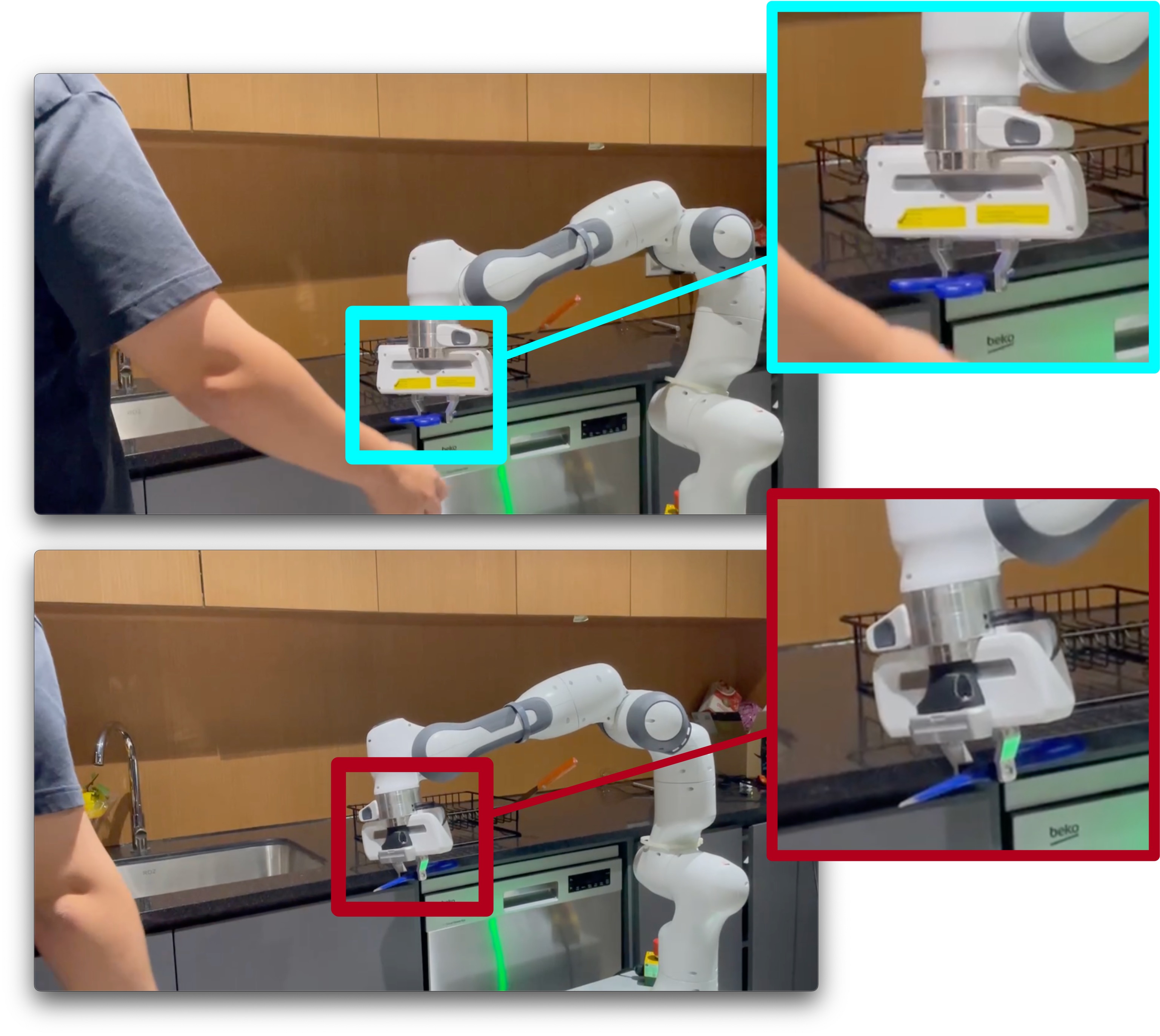}
    \caption{Success by passing the safe and clean handle (top) and intentional failure by passing the wrong end (bottom) while passing the scissors.}
    \label{fig:intentional_fail}
\vspace{-10pt}
\end{figure}

Participants are provided with rules similar to the table-clearing experiment: if the human stays put and the robot succeeds, they receive a reward of 1. Since the utensils are dirty, the handover is only considered successful if the robot passes the object in a manner that the human can easily grasp the clean handle.
If the human stays put and the robot fails, they receive a penalty of $-1$. If the human chooses to intervene, they will receive no reward or penalty. %

The following information is \emph{not} revealed to the participant: the robot is able to always succeed on the spatula, whisk, and scissors, but it will always fail to properly hand over the knife and drop it. When it happens, the experiment is terminated and a penalty of $-10$ is received. As such, the robot has to actively calibrate the human's trust so that the robot can help as much as possible (accumulating rewards in the process), yet influence the human to \textbf{not} trust it to pass the knife. {This setup is to simulate scenarios where humans may \emph{over-trust} robots' capability, thus requiring the robots to reduce human trust for better task performance over the long term}~\cite{chen2018planning}. Here, the robot intentionally fails to reduce trust, but future work can examine other strategies such as verbal communication. 
To facilitate this calibration process~\cite{Lee_Fong_Kok_Soh_2020}, we enable the robot to intentionally fail on all utensils (except the knife) by handing the wrong part to the human (so that the human can only grasp the dirty end, i.e. augment the robot's action space with these fail actions. Fig. \ref{fig:intentional_fail}). This intentional failure results in a penalty of $-1$.

\para{Hypothesis and Planners.} We consider two different plans: 
\begin{itemize}
    \item \textsc{LLM-Plan}, which is a deterministic plan generated by planning with a \textsc{davinci-TC-YN}-based human model with prompts similar to the table-clearing setup.;
    \item \textsc{Basic-Plan}, a myopic plan that ignores trust and always passes the spatula, egg whisk, scissors, and knife in that order (as the associated rewards are the same) and never intentionally fails, similar to the myopic plan used in~\cite{chen2018planning}. 
\end{itemize}
Our hypothesis was that a robot following \textsc{LLM-Plan} will reduce overtrust and yield higher returns compared to a robot following \textsc{Basic-Plan}.

\para{Participant Recruitment and Allocation.}
We recruited 65 participants from our university campus (ages 22 to 54). Participants were randomly divided into two groups. Each group was paired with either the \textsc{LLM-Plan} robot or the \textsc{Basic-Plan} robot. Due to safety considerations, participants did not physically interact with the robot. Instead, they completed an interactive video survey where a pre-recorded video of the robot's behavior was shown at each turn.

\para{Results.} 
21 out of the 33 participants ($63.6\%$) in the \textsc{Basic-Plan} group allowed the robot to pass the knife, compared to 9 out of 32 participants ($28.1\%$) in the \textsc{LLM-Plan} group. As a result, the mean return was higher for the \textsc{LLM-Plan} robot (-1.88) vs. \textsc{Basic-Plan} robot (-4.24), which a one-way ANOVA showed to be statistically significant at the $\alpha=5\%$ level ($F(1, 63) = [4.302]$, $p = 0.042$). These results support our hypothesis.

\para{Discussion.} 
Both \textsc{Basic-Plan} and \textsc{LLM-Plan} robots choose to pass the spatula and egg whisk in the first 2 turns. Then the plans diverge, depending on human actions. Unlike \textsc{Basic-Plan}, if the human decides not to intervene on the first two items, the \textsc{LLM-Plan} robot  chooses to intentionally fail on the scissors to prevent overtrust. A majority of participants who witnessed this intentional failure would then choose to intervene on the knife (18/20 participants, 90\%). As expected, those who experienced the \textsc{Basic-Plan} where the robot successfully handed over the scissors would \emph{not} intervene on the knife (1/15 participants intervened, 6.67\%).

However, not all participants experienced the above sequence of events. Five participants in the \textsc{LLM-Plan} group chose to intervene on the egg whisk in the second turn, which then prompted the robot to pass the knife in the next round (as it predicted the human would not trust it to pass the knife). However, these participants then chose \emph{not} to intervene. When asked to justify their decision, they explained that the spatula and knife were more similar due to their handle shape, compared to the egg whisk which had a shorter handle (Fig.~\ref{fig:utensil_passing}) and was perceived to be more difficult for the robot. A similar reason was given by the two participants who still chose to trust the robot to pass the knife after observing the robot fail to properly pass the scissors: the scissors was perceived to have a shape that was different from the other items. These comments suggest how perception of object shape plays a role in human-robot trust and present a challenge to incorporate such information into LLMs.

\section{Conclusion}
In this work, we present the first study into LLM-based zero-shot human models in HRI. Our key finding is that LLMs can be effective task-level human models --- they can model high-level human states and behavior. We further demonstrated that incorporating an LLM-based human model can yield reasonable plans in two trust-based social HRI scenarios. However, our research shows that current LLMs are unlikely to be accurate human models on their own; they have difficulty accounting for low-level geometrical/shape features due to limitations in spatial/numerical reasoning. Nevertheless, they can capture facets of human behavior, and combining LLMs with other models would make for interesting future research.

\bibliography{reference}

% Generated by IEEEtran.bst, version: 1.14 (2015/08/26)
\begin{thebibliography}{10}
\providecommand{\url}[1]{#1}
\csname url@samestyle\endcsname
\providecommand{\newblock}{\relax}
\providecommand{\bibinfo}[2]{#2}
\providecommand{\BIBentrySTDinterwordspacing}{\spaceskip=0pt\relax}
\providecommand{\BIBentryALTinterwordstretchfactor}{4}
\providecommand{\BIBentryALTinterwordspacing}{\spaceskip=\fontdimen2\font plus
\BIBentryALTinterwordstretchfactor\fontdimen3\font minus
  \fontdimen4\font\relax}
\providecommand{\BIBforeignlanguage}[2]{{%
\expandafter\ifx\csname l@#1\endcsname\relax
\typeout{** WARNING: IEEEtran.bst: No hyphenation pattern has been}%
\typeout{** loaded for the language `#1'. Using the pattern for}%
\typeout{** the default language instead.}%
\else
\language=\csname l@#1\endcsname
\fi
#2}}
\providecommand{\BIBdecl}{\relax}
\BIBdecl

\bibitem{choudhury2019utility}
R.~Choudhury, G.~Swamy, D.~Hadfield-Menell, and A.~D. Dragan, ``On the utility
  of model learning in hri,'' in \emph{2019 14th ACM/IEEE Intl. Conf. on
  Human-Robot Interaction (HRI)}.\hskip 1em plus 0.5em minus 0.4em\relax IEEE,
  2019, pp. 317--325.

\bibitem{ziebart2009planning}
B.~D. Ziebart, N.~Ratliff, G.~Gallagher, C.~Mertz, K.~Peterson, J.~A. Bagnell,
  M.~Hebert, A.~K. Dey, and S.~Srinivasa, ``Planning-based prediction for
  pedestrians,'' in \emph{2009 IEEE/RSJ Intl. Conf. on Intelligent Robots and
  Systems}.\hskip 1em plus 0.5em minus 0.4em\relax IEEE, 2009, pp. 3931--3936.

\bibitem{sadigh2016planning}
D.~Sadigh, S.~Sastry, S.~A. Seshia, and A.~D. Dragan, ``Planning for autonomous
  cars that leverage effects on human actions.'' in \emph{Robotics: Science and
  systems}, vol.~2.\hskip 1em plus 0.5em minus 0.4em\relax Ann Arbor, MI, USA,
  2016, pp. 1--9.

\bibitem{baker2017rational}
C.~L. Baker, J.~Jara-Ettinger, R.~Saxe, and J.~B. Tenenbaum, ``Rational
  quantitative attribution of beliefs, desires and percepts in human
  mentalizing,'' \emph{Nature Human Behaviour}, vol.~1, no.~4, p. 0064, 2017.

\bibitem{chen2018planning}
M.~Chen, S.~Nikolaidis, H.~Soh, D.~Hsu, and S.~Srinivasa, ``Planning with trust
  for human-robot collaboration,'' in \emph{Proceedings of the 2018 ACM/IEEE
  Intl. Conf. on human-robot interaction}, 2018, pp. 307--315.

\bibitem{chen2022mirror}
K.~Chen, J.~Fong, and H.~Soh, ``Mirror: Differentiable deep social projection
  for assistive human-robot communication,'' \emph{arXiv preprint
  arXiv:2203.02877}, 2022.

\bibitem{schmerling2018multimodal}
E.~Schmerling, K.~Leung, W.~Vollprecht, and M.~Pavone, ``Multimodal
  probabilistic model-based planning for human-robot interaction,'' in
  \emph{2018 IEEE Intl. Conf. on Robotics and Automation (ICRA)}.\hskip 1em
  plus 0.5em minus 0.4em\relax IEEE, 2018, pp. 3399--3406.

\bibitem{thomaz2016computational}
A.~Thomaz, G.~Hoffman, M.~Cakmak \emph{et~al.}, ``Computational human-robot
  interaction,'' \emph{Foundations and Trends{\textregistered} in Robotics},
  vol.~4, no. 2-3, pp. 105--223, 2016.

\bibitem{nocentini2019survey}
O.~Nocentini, L.~Fiorini, G.~Acerbi, A.~Sorrentino, G.~Mancioppi, and
  F.~Cavallo, ``A survey of behavioral models for social robots,''
  \emph{Robotics}, vol.~8, no.~3, p.~54, 2019.

\bibitem{kojima2022large}
T.~Kojima, S.~S. Gu, M.~Reid, Y.~Matsuo, and Y.~Iwasawa, ``Large language
  models are zero-shot reasoners,'' \emph{arXiv preprint arXiv:2205.11916},
  2022.

\bibitem{valmeekam2022large}
K.~Valmeekam, A.~Olmo, S.~Sreedharan, and S.~Kambhampati, ``Large language
  models still can't plan (a benchmark for llms on planning and reasoning about
  change),'' \emph{arXiv preprint arXiv:2206.10498}, 2022.

\bibitem{huang2022language}
W.~Huang, P.~Abbeel, D.~Pathak, and I.~Mordatch, ``Language models as zero-shot
  planners: Extracting actionable knowledge for embodied agents,'' in
  \emph{Intl. Conf. on Machine Learning}.\hskip 1em plus 0.5em minus
  0.4em\relax PMLR, 2022, pp. 9118--9147.

\bibitem{yaqi23translating}
\BIBentryALTinterwordspacing
Y.~Xie, C.~Yu, T.~Zhu, J.~Bai, Z.~Gong, and H.~Soh, ``Translating natural
  language to planning goals with large-language models,'' 2023. [Online].
  Available: \url{https://arxiv.org/abs/2302.05128}
\BIBentrySTDinterwordspacing

\bibitem{brown2020language}
T.~Brown, B.~Mann, N.~Ryder, M.~Subbiah, J.~D. Kaplan, P.~Dhariwal,
  A.~Neelakantan, P.~Shyam, G.~Sastry, A.~Askell \emph{et~al.}, ``Language
  models are few-shot learners,'' \emph{Advances in neural information
  processing systems}, vol.~33, pp. 1877--1901, 2020.

\bibitem{chowdhery2022palm}
A.~Chowdhery, S.~Narang, J.~Devlin, M.~Bosma, G.~Mishra, A.~Roberts, P.~Barham,
  H.~W. Chung, C.~Sutton, S.~Gehrmann \emph{et~al.}, ``Palm: Scaling language
  modeling with pathways,'' \emph{arXiv preprint arXiv:2204.02311}, 2022.

\bibitem{aher2022using}
G.~Aher, R.~I. Arriaga, and A.~T. Kalai, ``Using large language models to
  simulate multiple humans,'' \emph{arXiv preprint arXiv:2208.10264}, 2022.

\bibitem{kosinski2023theory}
M.~Kosinski, ``Theory of mind may have spontaneously emerged in large language
  models,'' \emph{arXiv preprint arXiv:2302.02083}, 2023.

\bibitem{ullman2023large}
T.~Ullman, ``Large language models fail on trivial alterations to
  theory-of-mind tasks,'' \emph{arXiv preprint arXiv:2302.08399}, 2023.

\bibitem{ahn2022can}
M.~Ahn, A.~Brohan, N.~Brown, Y.~Chebotar, O.~Cortes, B.~David, C.~Finn,
  K.~Gopalakrishnan, K.~Hausman, A.~Herzog \emph{et~al.}, ``Do as i can, not as
  i say: Grounding language in robotic affordances,'' \emph{arXiv preprint
  arXiv:2204.01691}, 2022.

\bibitem{lee2019bayesian}
J.~J. Lee, F.~Sha, and C.~Breazeal, ``A bayesian theory of mind approach to
  nonverbal communication,'' in \emph{2019 14th ACM/IEEE Intl. Conf. on
  Human-Robot Interaction (HRI)}.\hskip 1em plus 0.5em minus 0.4em\relax IEEE,
  2019, pp. 487--496.

\bibitem{baker2014modeling}
C.~L. Baker and J.~B. Tenenbaum, ``Modeling human plan recognition using
  bayesian theory of mind,'' \emph{Plan, activity, and intent recognition:
  Theory and practice}, vol.~7, pp. 177--204, 2014.

\bibitem{strouse2021collaborating}
D.~Strouse, K.~McKee, M.~Botvinick, E.~Hughes, and R.~Everett, ``Collaborating
  with humans without human data,'' \emph{Advances in Neural Information
  Processing Systems}, vol.~34, pp. 14\,502--14\,515, 2021.

\bibitem{tjomsland2022mind}
J.~Tjomsland, S.~Kalkan, and H.~Gunes, ``Mind your manners! a dataset and a
  continual learning approach for assessing social appropriateness of robot
  actions,'' \emph{Frontiers in Robotics and AI}, vol.~9, p.~4, 2022.

\bibitem{soh2018transfer}
H.~Soh, P.~Shu, M.~Chen, and D.~Hsu, ``The transfer of human trust in robot
  capabilities across tasks.'' in \emph{R:SS}, 2018.

\bibitem{soh2020multi}
H.~Soh, Y.~Xie, M.~Chen, and D.~Hsu, ``Multi-task trust transfer for
  human--robot interaction,'' \emph{The Intl. Journal of Robotics Research},
  vol.~39, no. 2-3, pp. 233--249, 2020.

\bibitem{sap2019socialiqa}
M.~Sap, H.~Rashkin, D.~Chen, R.~LeBras, and Y.~Choi, ``Socialiqa: Commonsense
  reasoning about social interactions,'' \emph{arXiv preprint
  arXiv:1904.09728}, 2019.

\bibitem{ouyang2022training}
L.~Ouyang, J.~Wu, X.~Jiang, D.~Almeida, C.~L. Wainwright, P.~Mishkin, C.~Zhang,
  S.~Agarwal, K.~Slama, A.~Ray \emph{et~al.}, ``Training language models to
  follow instructions with human feedback,'' \emph{arXiv preprint
  arXiv:2203.02155}, 2022.

\bibitem{chung2022scaling}
H.~W. Chung, L.~Hou, S.~Longpre, B.~Zoph, Y.~Tay, W.~Fedus, E.~Li, X.~Wang,
  M.~Dehghani, S.~Brahma \emph{et~al.}, ``Scaling instruction-finetuned
  language models,'' \emph{arXiv preprint arXiv:2210.11416}, 2022.

\bibitem{suppmat}
B.~Zhang and H.~Soh, ``{Supplementary Material for Large Language Models as
  Zero-Shot Human Models for Human-Robot Interaction},''
  \url{https://github.com/clear-nus/llm-human-model}, 2023.

\bibitem{devlin2018bert}
J.~Devlin, M.-W. Chang, K.~Lee, and K.~Toutanova, ``Bert: Pre-training of deep
  bidirectional transformers for language understanding,'' \emph{arXiv preprint
  arXiv:1810.04805}, 2018.

\bibitem{heyes2014cultural}
C.~M. Heyes and C.~D. Frith, ``The cultural evolution of mind reading,''
  \emph{Science}, vol. 344, no. 6190, p. 1243091, 2014.

\bibitem{zhang2012perspective}
J.~Zhang, T.~Hedden, and A.~Chia, ``Perspective-taking and depth of
  theory-of-mind reasoning in sequential-move games,'' \emph{Cognitive
  science}, vol.~36, no.~3, pp. 560--573, 2012.

\bibitem{srivastava2022beyond}
A.~Srivastava, A.~Rastogi, A.~Rao, A.~A.~M. Shoeb, A.~Abid, A.~Fisch, A.~R.
  Brown, A.~Santoro, A.~Gupta, A.~Garriga-Alonso \emph{et~al.}, ``Beyond the
  imitation game: Quantifying and extrapolating the capabilities of language
  models,'' \emph{arXiv preprint arXiv:2206.04615}, 2022.

\bibitem{Lee_Fong_Kok_Soh_2020}
J.~Lee, J.~Fong, B.~C. Kok, and H.~Soh, ``Getting to know one another:
  Calibrating intent, capabilities and trust for human-robot collaboration,''
  \emph{IEEE Intl. Conf. on Intelligent Robots and Systems}, p. 6296–6303,
  2020.

\end{thebibliography}
\bibliographystyle{IEEEtran}

\clearpage

\end{document}